%% file: manuscript.tex
\documentclass[sigconf]{acmart}

\AtBeginDocument{%
  \providecommand\BibTeX{{%
    \normalfont B\kern-0.5em{\scshape i\kern-0.25em b}\kern-0.8em\TeX}}}

\setcopyright{acmcopyright}
\copyrightyear{}
\acmYear{}
\acmDOI{}

\acmConference[]{}{}{}
\acmBooktitle{}
\acmPrice{}
\acmISBN{}

\begin{document}

\title{A psychophysics approach for quantitative comparison of interpretable computer vision models}


\author{Felix Biessmann}
\email{felix.biessmann@beuth-hochschule.de}
\affiliation{%
  \institution{Beuth University}
  \city{Berlin}
  \country{Germany}
}

\author{Dionysius Refiano}
\email{s73108@beuth-hochschule.de}
\affiliation{%
  \institution{Beuth University}
  \city{Berlin}
  \country{Germany}
}

\renewcommand{\shortauthors}{Biessmann and Refiano}

\begin{abstract}
The field of transparent Machine Learning (ML) has  contributed many novel methods aiming at better interpretability for computer vision and ML models in general. But how useful the explanations provided by transparent ML methods are for humans remains difficult to assess. Most studies evaluate interpretability in qualitative comparisons, they use experimental paradigms that do not allow for direct comparisons amongst methods or they report only offline experiments with no humans in the loop. While there are clear advantages of evaluations with no humans in the loop, such as scalability, reproducibility and less algorithmic bias than with humans in the loop, these metrics are limited in their usefulness if we do not understand how they relate to other metrics that take human cognition into account. 
Here we investigate the quality of interpretable computer vision algorithms using techniques from psychophysics. In crowdsourced annotation tasks we study the impact of different interpretability approaches on annotation accuracy and task time. In order to relate these findings to quality measures for interpretability without humans in the loop we compare quality metrics with and without humans in the loop. Our results demonstrate that psychophysical experiments allow for robust quality assessment of transparency in machine learning. Interestingly the quality metrics computed without humans in the loop did not provide a consistent ranking of interpretability methods nor were they representative for how useful an explanation was for humans. 
These findings highlight the potential of methods from classical psychophysics for modern machine learning applications. We hope that our results provide convincing arguments for evaluating interpretability in its natural habitat, human-ML interaction, if the goal is to obtain an authentic assessment of interpretability.
\end{abstract}

%



\maketitle

\input{intro}
\input{related_work}
\input{experiments}
\input{results}
\input{conclusion}


\bibliographystyle{ACM-Reference-Format}
\bibliography{references}

%

\end{document}

%% file: intro.tex

\section{Introduction}
\label{sec:intro}

In recent years complex machine learning (ML) models, many based on deep learning, have achieved surprising results in computer vision, natural language processing and many other domains. These models are difficult to interpret, which inspired many researchers to investigate ways to render ML models {\em interpretable}~\cite{Kim2015,lipton2016mythos,doshi2017towards,Herman2017ThePA}. There are many motivations for interpretable ML methods. Domain experts, data scientists or data engineers that control proper functioning of an ML pipeline need to be be able to access the rules learned by a ML system in an intuitive manner in order to quickly spot the root causes of errors. More generally the main motivation for research on transparent ML is that intuitive human understanding of ML predictions can is a prerequisite for a healthy trust relationship between humans and assistive ML systems. In particular transparency is argued to prevent algorithm aversion as well as algorithmic bias. Algorithm aversion refers to cases when humans do not trust ML systems, even when they know that the model predictions are more accurate than those of a human \cite{Dietvorst2015}, algorithmic bias are cases of ethnical or gender biases in ML predictions \cite{Hajian2016}. In the following we will also use the term algorithmic bias to refer to cases of too much trust into an ML prediction, for instance when a human interacting with assistive ML technology blindly follows its predictions. The usual narrative is that explanations of ML decisions can increase human trust in them \cite{Sinha2002, lime}. 
%

A central problem with interpretability methods is that they are difficult to compare and evaluate. Most of the research  compares methods using either proxy measures, that do not directly relate to interpretability by humans, as e.g. \cite{Samek2017}, or qualitative measures that render comparisons of results across studies difficult \cite{Strumbel2010}. In this work we propose to use psychophysical methods to quantify and compare the quality of interpretability methods.
We follow the ideas of \cite{schmidt2019quantifying} and base our approach on the assumption that the definition of interpretability is inherently tied to a human observer. Good interpretability methods should allow human observers to intuitively understand a ML prediction. Intuitive understanding of the rules learned by a ML system is reflected in how accurately and how fast humans make decisions when assisted with a transparent ML prediction. These two variables can be easily measured in psychophysical experiments that study the interaction between humans and ML systems. 

The motivation for this work is twofold: For one this work aims at complementing previous work on measuring the quality of interpretability methods by establishing a quantitative measure of interpretability in the domain of computer vision that captures aspects of human cognition. Ultimately this will help practitioners to choose the right interpretability method for a given use case and researchers to devise novel objectives for better interpretability methods. Secondly the goal of this study is to validate to what extent existing approaches for measuring interpretability without humans in the loop reflect the interpretability metrics we measure in psychophysical experiments. 

In the following we shortly highlight some of the related work and then describe an image annotation task, emotion recognition, as well as the ML model, the transparency approaches used and the experimental design for quantitatively evaluating interpretability with humans in the loop (HIL) and with no humans in the loop (NHIL). We compare the different interpretability approaches with respect to the HIL and NHIL metrics and analyze their relationship, in particular whether cheaper and more scalable machine based NHIL transparency metrics reflect the most relevant but more expensive HIL transparency metrics. We conclude with highlighting the implications of our results for practitioners that build systems with human-ML interaction or transparent ML. 

%% file: related_work.tex

\section{Related Work}
\label{sec:relatedwork}

While the literature on evaluation on transparent ML is very diverse \cite{lipton2016mythos}, there appears to be a consensus in the literature that model explanations should overlap with human intuitions and that there is a lack of quantitative evaluation standards \cite{miller2017explanation,doshi2017towards,lipton2017doctor}.
The technical contributions to the field of transparent ML can be broadly categorized into two types of methods. First there are methods that aim at rendering specific models interpretable, such as interpretability methods for linear models \cite{haufe2014interpretation} or interpretability for neural network models \cite{Zeiler2014,Simonyan2013,Montavon2017}. Second there are interpretability approaches that aim at rendering {\em any} model interpretable, a popular example are the {\em Local Interpretable Model-Agnostic Explanations} (LIME) \cite{lime}. As these latter interpretability methods do not need to have access to the inner workings of a ML model, they are often referred to as {\em black box interpretability methods}.
One of the challenges with most interpretability approaches is that it is difficult to evaluate how interpretable to humans a model prediction becomes when employing a given interpretability method. 
The most straightforward approach to evaluation of interpretability is to generate synthetic data from a known generative model and evaluate the explanations against the true data generation process  \cite{Zien2009,haufe2014interpretation}. However it can be very challenging to design generative models for real data. 

In the field of computer vision there have been a number of interpretability approaches specialized for that application scenario and the method of choice in this field, deep neural networks. Some prominent examples include {\em layerwise relevance propagation} (LRP) \cite{Lapuschkin2017}, sensitivity analysis \cite{Simonyan2013} and deconvolutions \cite{Zeiler2014}. For comparing these different approaches the authors of \cite{Samek2017} propose a greedy iterative perturbation procedure for comparing LRP, sensitivity analysis and deconvolutions. The idea is to remove features where the perturbation probability is proportional to the relevance score of each feature given by the respective interpretability method. 
The idea of using perturbations underlies also many other interpretability approaches, such as the work on {\em influence functions} \cite{Cook1977,pmlr-v70-koh17a,hampel2011robust} and methods based on game theoretic insights \cite{Strumbel2010,Lundberg2017}. 

While there are comprehensive surveys on this matter \cite{Guidotti2018}, the evaluation criteria are often problematic and in many cases do not allow a direct comparison of methods. 
Most attempts to evaluate interpretability methods either rely on proxy measures that are not related directly to interpretability, such as runtime or robustness of the interpretability model under perturbations or they merely use  qualitative measures, as in e.g. \cite{Strumbel2010}. 
Reflecting the intuition that interpretability cannot be evaluated without taking a human in the loop there is increasing interest in investigating the quality of transparent ML methods in psychological experiments on human-machine interaction \cite{HUYSMANS2011,Lundberg2017,ribeiro2018anchors,lakkaraju2016interpretable,schmidt2019quantifying}.

Building on these results we here employ psychophysical experiments in a crowdsourcing scenario in order to evaluate the quality of interpretability methods. This quality measure is closely related to the approach taken in previous work \cite{lakkaraju2016interpretable,schmidt2019quantifying} but we here focus on the computer vision domain, which requires specific experimental designs as visual cognition is very different from cognition of text and semantics. For instance visual cognition is characterized by much faster processing speed compared to text understanding as done in \cite{lakkaraju2016interpretable,schmidt2019quantifying}.
One aspect that is, to the best of our knowledge, underrepresented in the field of transparent computer vision algorithms is a comprehensive comparison between human in the loop metrics and more efficient machine based metrics. Without an in depth understanding of how machine based metrics relate to metrics that capture human cognition, it is difficult to assess the true quality of an interpretability method that was evaluated with machine based metrics only. 

%% file: experiments.tex

\section{Experiments}
\label{sec:experiments}
In the following we describe the annotation task and the technical prerequisites of our experiments, including the ML model used and the transparency approaches applied to it. We then explain the experimental paradigm for both interpretability evaluation with psychophysical experiments as well as the more commonly used evaluation with no humans in the loop.

\subsection{Annotation Task}
\begin{figure}
\includegraphics[width=\columnwidth]{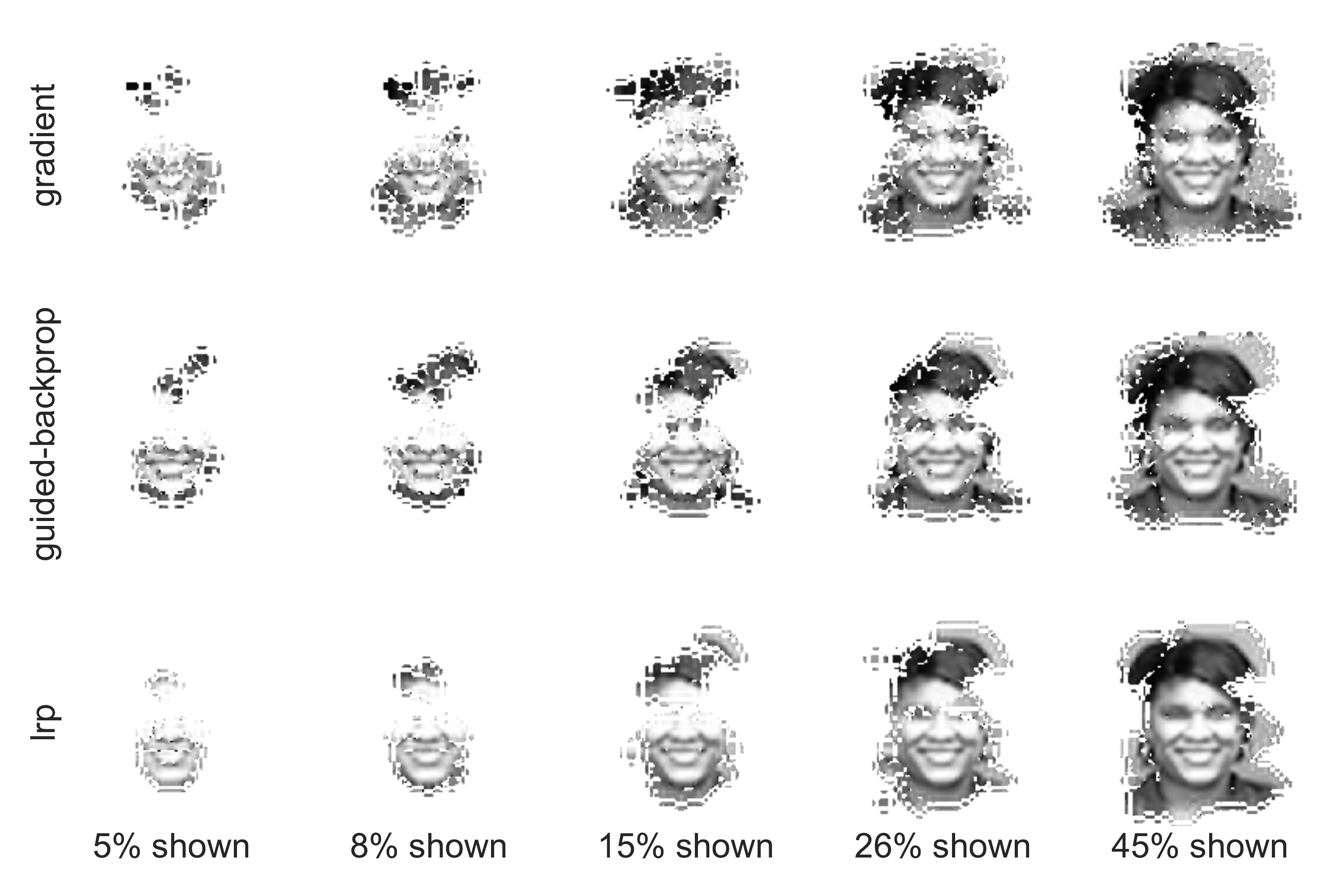}\\
\includegraphics[width=\columnwidth]{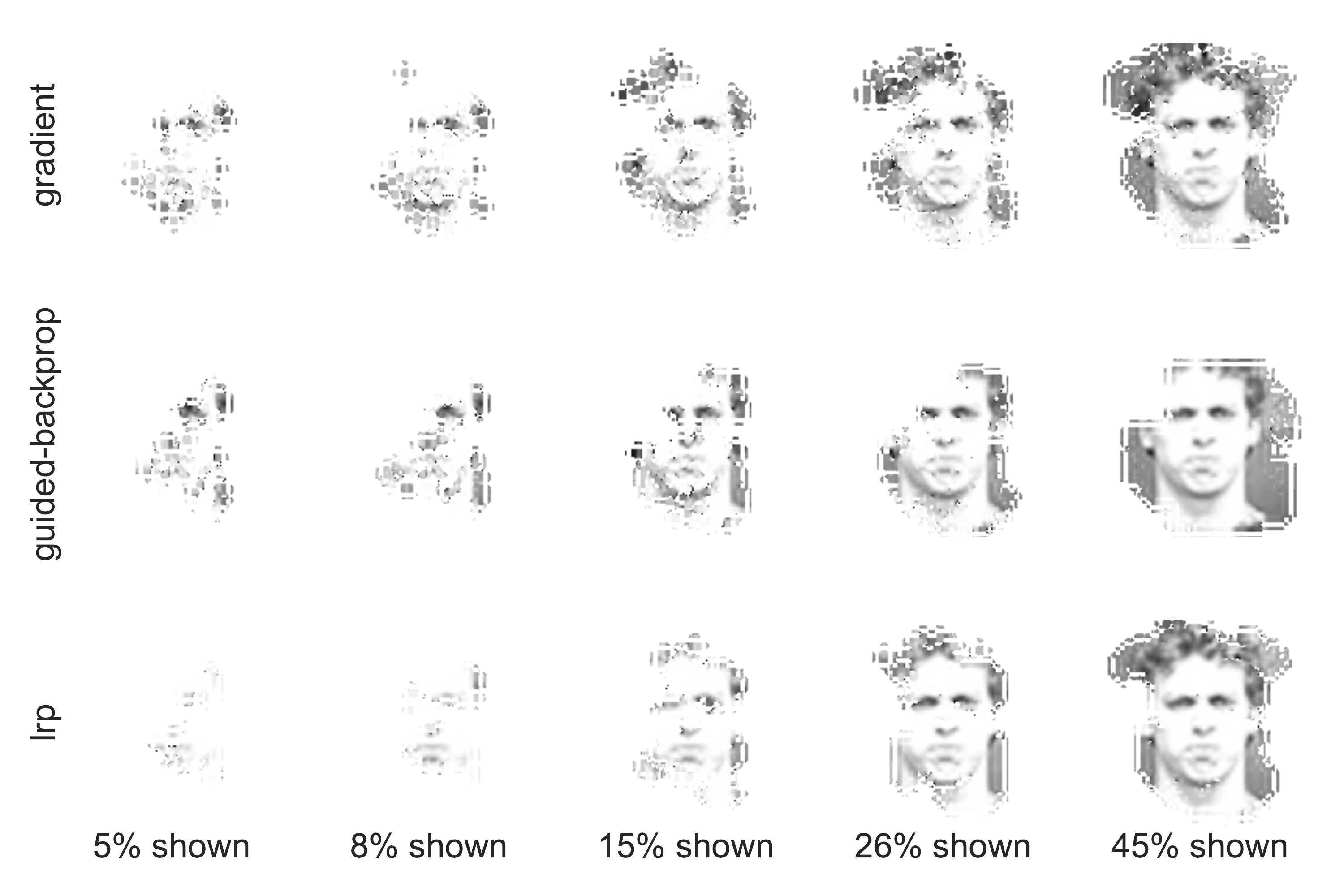}
\caption{Examples of masked images with happy emotional expression (\textit{top three rows}) and angry expression (\textit{bottom three rows}) for different interpretability methods (gradient, guided-backprop \cite{Springenberg2014} and LRP \cite{Lapuschkin2017}, shown in rows) and mask sizes (shown in columns)} 
\label{fig:example}
\end{figure}

The annotation task was emotional expression classification on images. We used the extended Cohn-Kanade image data set \cite{Lucey2010} which contains images for the classes, anger, contempt, disgust, fear, happiness, sadness, surprise. The class distribution can be seen in \autoref{tab:eckcd}. We reduced the data to a binary classification task in which annotators had to classify emotional expressions of {\em anger} and {\em happiness}. Some sample images are shown in \autoref{fig:example}. The annotators had the option of not providing an annotation in case they did not recognize the emotional expression. 
We chose this data set over other standard benchmark tasks in the domain of computer vision as it did not involve the localization of the target object but rather the detection of a complex pattern in human faces. When applying interpretability methods to models applied to other benchmarks, like ImageNet \cite{imagenet_cvpr09}, the explanations computed often focus on localization of the target object. This effect can be considered a convenient proxy for determining whether the model has learned the right features; for instance if the model explanation correctly localizes the target object, this is better than when the model explanation focuses on features that are not the target object but just correlate with its appearance in the training data set. An example of the latter would be a model explanation that highlights the basketball court when it should focus on the basketball to predict the target object basketball. This form of overfitting is not unusual in models trained on common benchmark data sets and can be detected with interpretability methods. But we felt that for our purposes this is a confounding factor when we are interested in interpretability quality; we hence opted for the emotional expression task which did not require localization of the target object. 

\begin{table}[]
\begin{tabular}{lr}
\toprule
class &  number of images \\
\midrule
anger &                45 \\
contempt &                18 \\
disgust &                59 \\
fear &                25 \\
happiness &                69 \\
sadness &                28 \\
surprise &                83 \\
\bottomrule
\end{tabular}        
\caption{Class distribution of extended Cohn-Kanade (CK+) data set. We binarized the data set by extracting only images from the anger and happiness class.}
\label{tab:eckcd}
\end{table}
    
\subsection{Machine Learning Model}
For our experiments we used a computer vision model from an open source python toolkit that achieves state of the art performance on emotional expression prediction from images \cite{emopy}. We used the default model without any modifications. The precision, recall, and F1 scores on the data set used in our experiments are shown in \autoref{tab:emopyscore}. Note that these predictive performances are not perfect, but they can be considered competitive with the state of the art for this particular classification task. We also emphasize that we here are focussing on interpretability methods, not the underlying ML model. The quality of the machine learning model was the same for all interpretability methods. 

\begin{table}[]
\begin{tabular}{llllllll}
\toprule
emotion & precision & recall & f1-score & support  \\
\midrule
anger & 0.48 & 0.47 & 0.47 & 45 \\
happiness & 0.66 & 0.67 & 0.66 & 69 \\
\midrule
avg / total & 0.59 & 0.59 & 0.59 & 114 \\ 
\bottomrule
\end{tabular}
\caption{Held-out per label precision/recall/f1 scores of EmoPy used for comparing ML interpretability methods on the CK+ dataset}
\label{tab:emopyscore}
\end{table}

\subsection{Interpretability Methods}
We compared three different interpretability methods for the EmoPy computer vision model
\begin{itemize}
\item {\em Gradient}: The gradient of the output w.r.t. the input image
\item {\em Layerwise relevance propagation (lrp)}: attributes importance recursively to each neuron's input relevance \cite{Lapuschkin2017}
\item {\em Guided backpropagation}: applies ReLU in gradient computation in addition to the gradient of a ReLU \cite{Springenberg2014}
\end{itemize}
For all methods we used the implementation in the \texttt{iNNvestigate!} package \cite{Alber2018}. All methods were used with their default hyperparameters. For the LRP approach we used the \texttt{sequential\_preset\_a} variant provided in the package. 
The list of methods is not meant to be exhaustive. The main purpose of this work is to illustrate that the combination of psychophysical methods and ML can be helpful for quantifying the usefulness of interpretability methods. For the sake of simplicity, we deliberately restricted the set of interpretability methods to just three methods that other experts in the field recommended to us as useful. 

\subsection{Quality of Explanations}
\label{sec:quality_metrics}
We employ two different metrics to compare the quality of interpretability approaches, one standard approach similar to the commonly used quality metrics with no humans in the loop (NHIL) and one approach based on psychophysical experiments with humans in the loop (HIL). In both settings we use all three interpretability methods to compute scores for each pixel in the image. These scores roughly speaking capture the importance of that pixel for the model's prediction. Based on these scores we rank the pixels and mask a certain percentage of pixels. The percentages of shown pixels were ten logarithmically spaced values between 0 and 100 to account for the Weber-Fechner law postulating a logarithmic relationship between stimulus and perception \cite{Fechner1860}. The masks showed 5,  6,  8, 11, 15, 19, 26, 34, 45, 60 percent of pixels of the image. Some example images for the emotional expression {\em anger} and {\em happiness} are shown in \autoref{fig:example}. These thresholds were based on initial experiments with different thresholds in which we determined the minimum number of pixels needed to detect the emotion and the number of pixels needed to enable most subjects to correctly classify the image. 

\paragraph{No humans in the loop (NHIL) metrics}
When developing a new interpretability approach it is most convenient for researchers to iterate quickly on model improvements and to validate the improvements with tests that are ideally fast and can be conducted without humans in the loop. Most of these NHIL metrics perturb the input data in some way that takes into account the feature scores provided by an interpretability method. For instance in \cite{Samek2017} the authors replace small patches in an input image with noise and evaluate the predictive performance for each perturbation of the data.  We follow this idea and slightly modify the perturbations to match the conditions used in the psychophysics experiments. In particular we mask a certain percentage of pixels and feed the masked image to the convolutional neural network to obtain a prediction. To evaluate the interpretability quality we evaluate the predictive performance of the EmoPy model on masked images.

\paragraph{Psychophysical human in the loop (HIL) metrics}
In order to quantify the quality of interpretability methods in HIL psychophysical experiments we adopt the ideas from \cite{schmidt2019quantifying}, 
\begin{enumerate}
\item Interpretability is associated with {\em intuitive understanding}
\item Intuitive understanding leads to fast and accurate decisions
\end{enumerate}
Accuracy and speed of AI-assisted decisions can give insights into the cognitive load inherent to understanding of ML predictions. When an explanation is intuitive we will follow it without too much thinking; but when we need more time to digest an explanation, its relative interpretability quality is lower compared to other explanations. More importantly, when ML assisted decisions are followed quickly even in cases when the ML predictions were wrong, this is a clear sign of unhealthy algorithmic bias. Evaluating both, reaction time and accuracy of annotations, can thus provide authentic and quantifiable metrics of interpretability quality. 
Based on these ideas we measured the annotation accuracy as well as reaction times in the above emotional expression classification task. In the experiments we systematically controlled the amount of pixels unmasked by a given interpretability method to investigate the dependency of the signal strength and the interpretability. 

\subsection{User Interface and Experimental Design}
\begin{figure}
\includegraphics[width=.7\columnwidth]{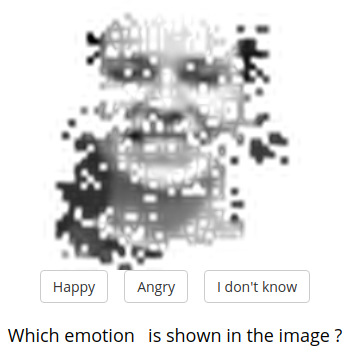}
\caption{User interface of a stimulus shown in the experiment. The image stimulus for a given mask size is located in the center of the page. Below the image buttons for annotating the image with the emotional expression \textit{anger} and \textit{happiness} are shown, along with an \textit{I don't know} option.} 
\label{fig:experimentinst}
\end{figure}

We built the user interface using the open source library jsPsych \cite{deleeuw2015}. The library provides basic features to design a psychological experiment running in the browser. In our case, we used the package to build an experiment timeline that showed the image stimulus with an html button below to capture the annotation provided by the experimental subjects.  
In each trial of an experiment we show the same image with increasing percentages of pixels shown. As we used ten different mask sizes from 5\% to 60\% of all pixels in the image, subjects saw a series of ten images. For illustration we show a subset of those ten masks for each interpretability method in \autoref{fig:example}. At the last image, when 60\% of the image was shown, all subjects correctly identified the emotional expression, see also \autoref{fig:accuracy}. 
The entire experiment was designed to be completed in about 10 minutes, based on a pilot experiment. For each label five images were shown, which resulted in $5~\text{(images)}~\times 2~\text{(classes)}~\times 3 ~\text{(interpretability methods)} \times 10~\text{(mask sizes)} = 300$ images in total that were annotated by each subject. 
The image stimulus is displayed in the format as shown in \autoref{fig:experimentinst}. 
For each subject the order of the trials was randomized, so each subject has seen each interpretability method and source image in a random order, but the order of unmasking the image was always the same. In total 62 subjects participated in the experiment.

The experiments were conducted on the crowdsourcing platform Amazon Mechanical Turk. We payed all subjects the minimum wage in the country of the research institution of the authors, 11\$US per hour. Mechanical Turk requires to show a preview of the experiment, before the worker accepts to participate. For the preview, we provided an instruction and an example trial.  
In the main part of the experiment, after the workers agreed to participate, they will be first shown the number of trials they need to complete. When they proceed, the actual experiment will start and will be completed after the subjects have annotated 300 images. 

%% file: results.tex

\section{Results}
\label{sec:results}

In the following we first analyse the results of the psychophysical experiments and the results from the experiments without humans in the loop independently. Then we compare the interpretability metrics from both approaches. Lastly we also investigate the impact of different transparency approaches to negative forms of algorithmic bias. 

\begin{figure}
\includegraphics[width=7cm]{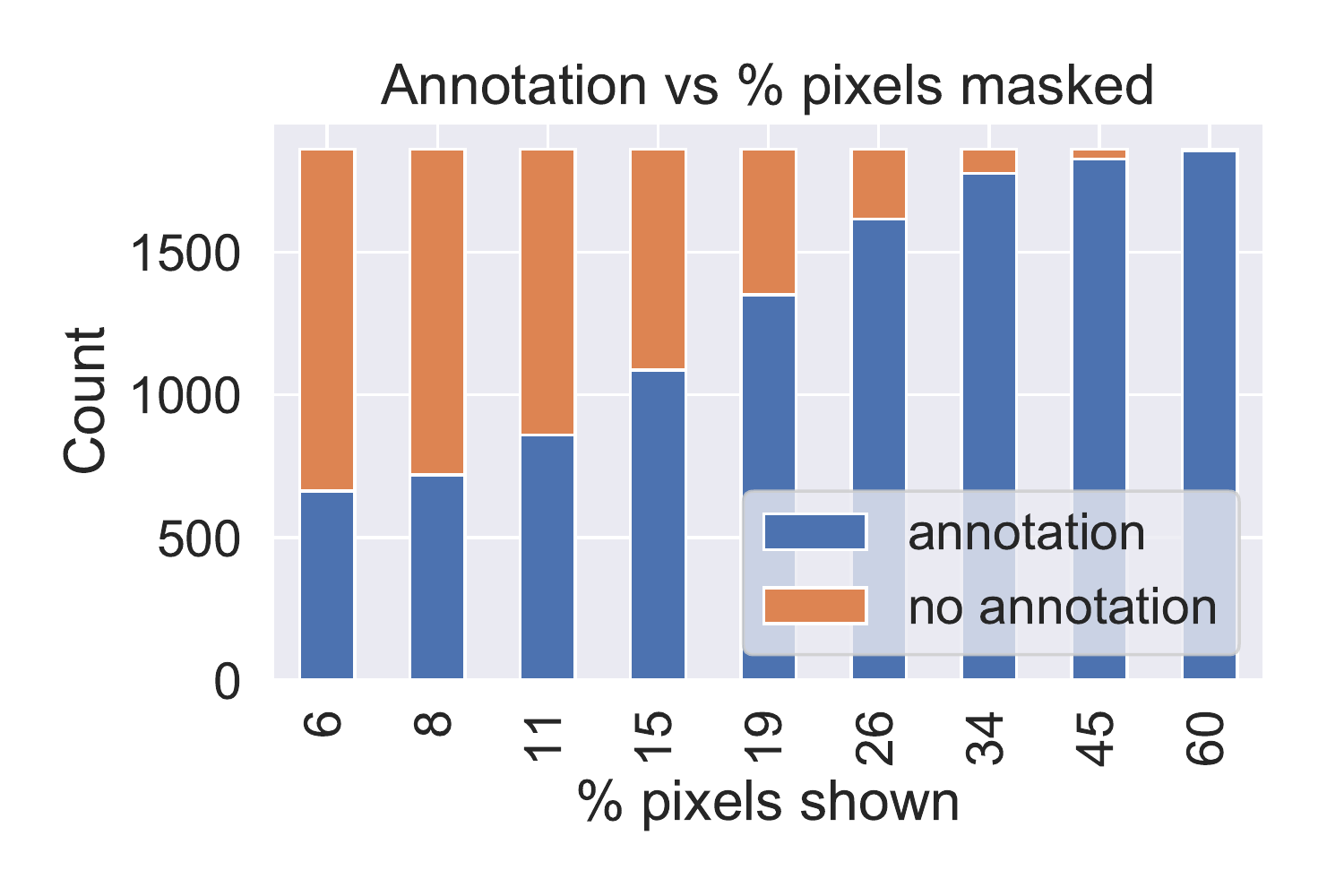}
\caption{Annotators' confidence, measured by counting how often they did not provide a label but the \textit{I don't know} label, as function of the mask size, aggregated over all interpretability methods. When 6\% of all pixels were shown, 663 annotators detected an emotion and provided a label, while 1197 annotators did not detect an emotion and provided only the \textit{I don't know} label. When 60\% of pixels were shown, all annotators detected the emotion.}
\label{fig:uncertainty_all}
\end{figure}

\begin{figure}
\includegraphics[width=8.5cm]{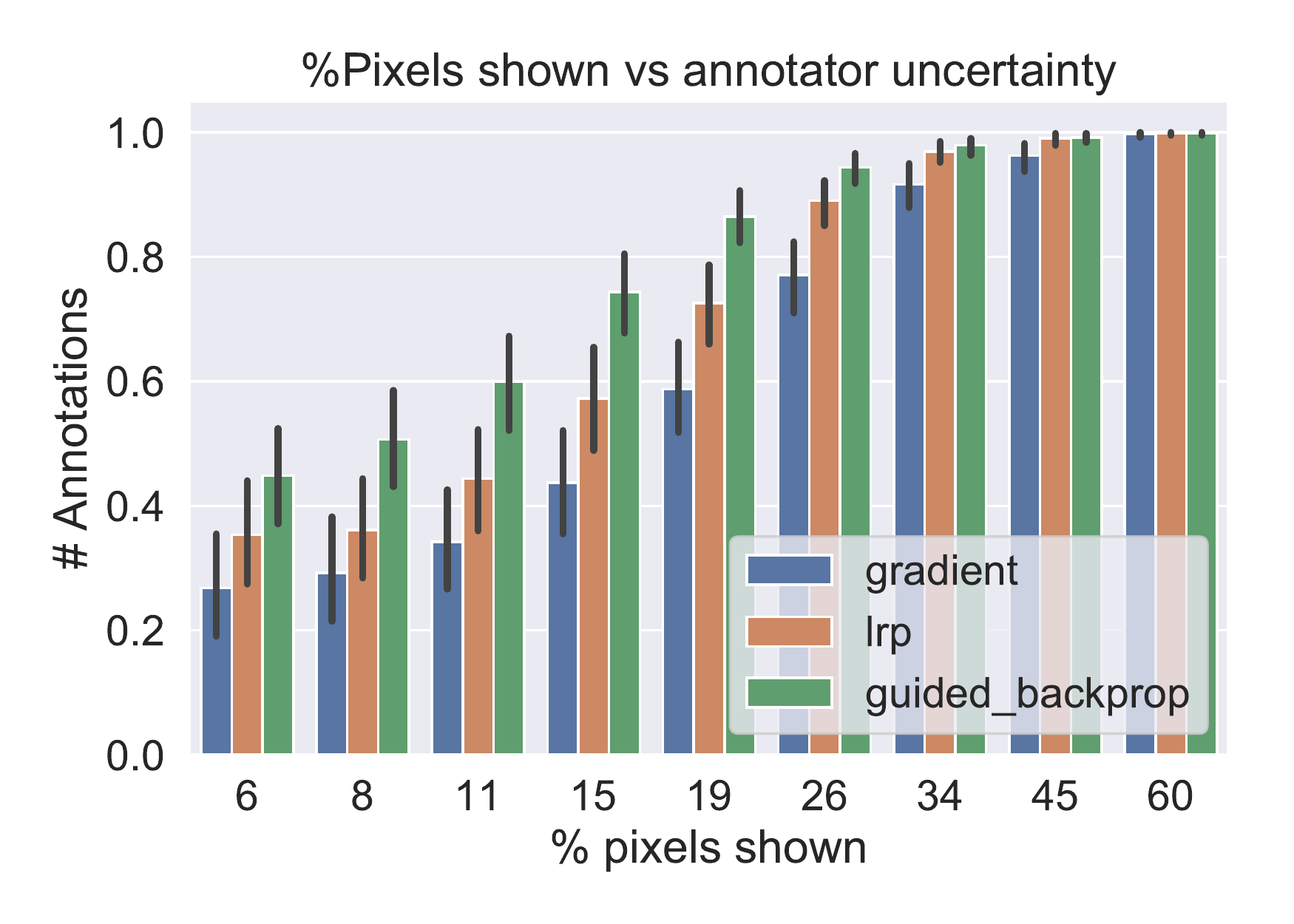}
\caption{Annotators' confidence, measured by counting how often they did not provide a label but the \textit{I don't know} label, as function of the mask size, for each interpretability methods. When only 6\% to 15\% of all pixels are shown, Guided BackProp assisted annotators are almost twice as certain as annotators assisted by Gradient explanations and provide annotations instead of the \textit{I don't know} label.}
\label{fig:uncertainty}
\end{figure}

\subsection{Psychophysical Experiments}
\label{sec:psychophysics_results}

\paragraph{Annotators' uncertainty and interpretability}
We investigated the impact of each interpretability approach on the uncertainty of annotators by counting how often they did not provide an annotation but just the \textit{I don't know} label. Averaging across all interpretability approaches we see in \autoref{fig:uncertainty_all} that the experimental settings were chosen such that there is a smooth increase in annotators' confidence when increasing the percentage of pixels of an image. Splitting the data into the different interpretability conditions, there is a clear effect of the interpretability method as shown in \autoref{fig:uncertainty}. The simplest gradient approach leads to the highest annotator uncertainty and least number of annotations up to mask sizes of 45\% of all pixels. Guided BackProp \cite{Springenberg2014} in contrast leads consistently to the lowest annotator uncertainty and the highest number of annotations. Comparing Gradient and Guided BackProp we find that on average almost twice as many annotators are certain enough about their prediction that they provide an annotation when assisted with the Guided BackProp saliency map, compared to the Gradient explanation that more often led annotators to choose the \textit{I don't know} label. This finding highlights the importance of quantitatively comparing transparency approaches. The extent to which human users of ML can profit from transparency strongly depends on the quality of the explanation provided. 

\begin{figure}
\includegraphics[width=8.5cm]{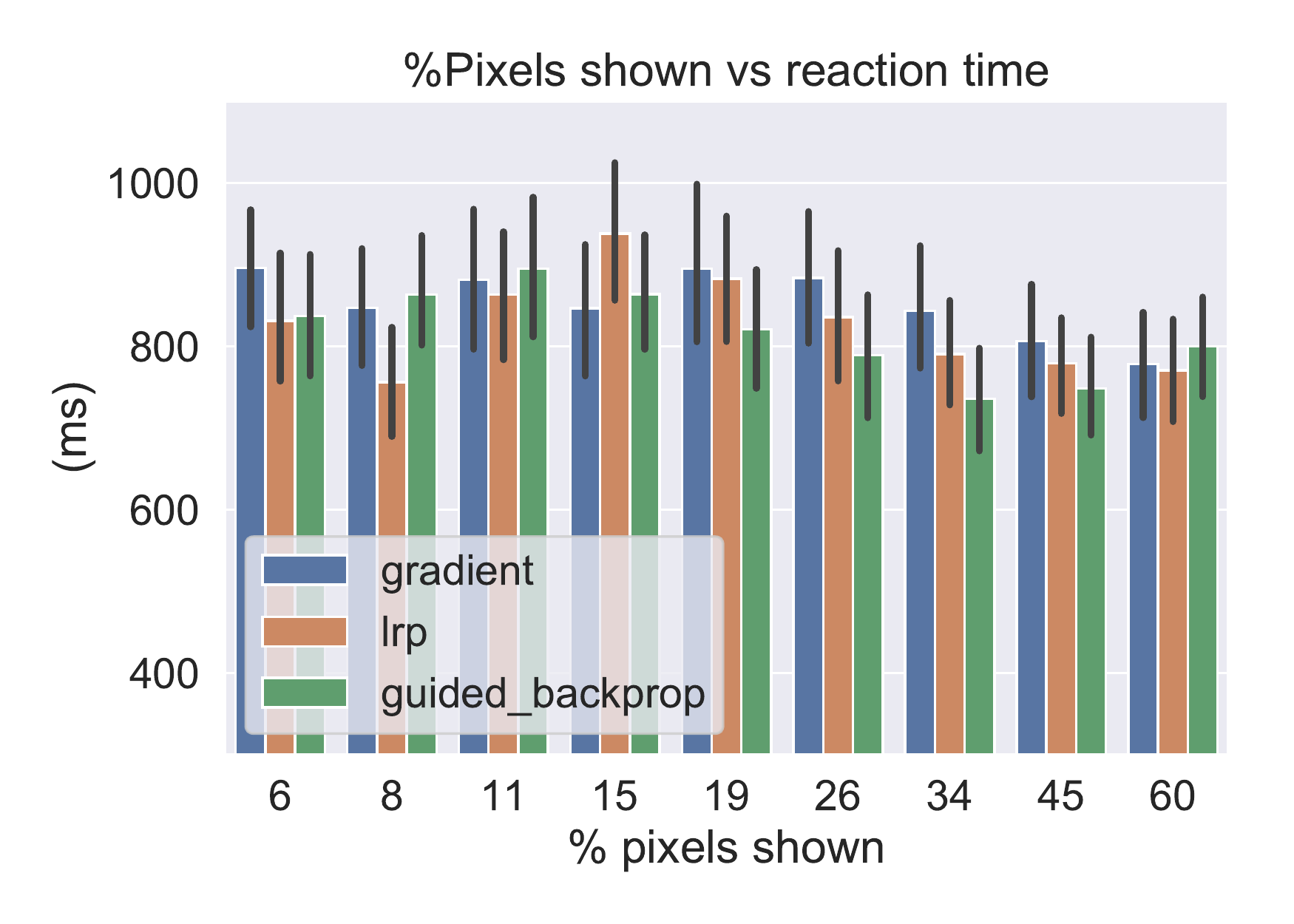}
\caption{Reaction times of all subjects for different interpretability methods and mask sizes. 
Cognitive load related to processing the explanation appears to peak around mask sizes of 15\% pixels and decays for larger masks once annotators have detected the emotional expression shown.}
\label{fig:rt}
\end{figure}

\paragraph{Reaction times reflect cognitive load of interpretatons}
In \autoref{fig:rt} we show the reaction times for each experimental condition. 
When most pixels are masked reaction times are low, as most subjects understand that they cannot make a correct prediction. For intermediate mask sizes around 15\% pixels shown, reaction times show a slight increase reflecting the increased cognitive load. For larger mask sizes, the reaction time decreases, as most subjects have provided an annotation already and keep clicking that label.

\begin{figure}
\includegraphics[width=8.5cm]{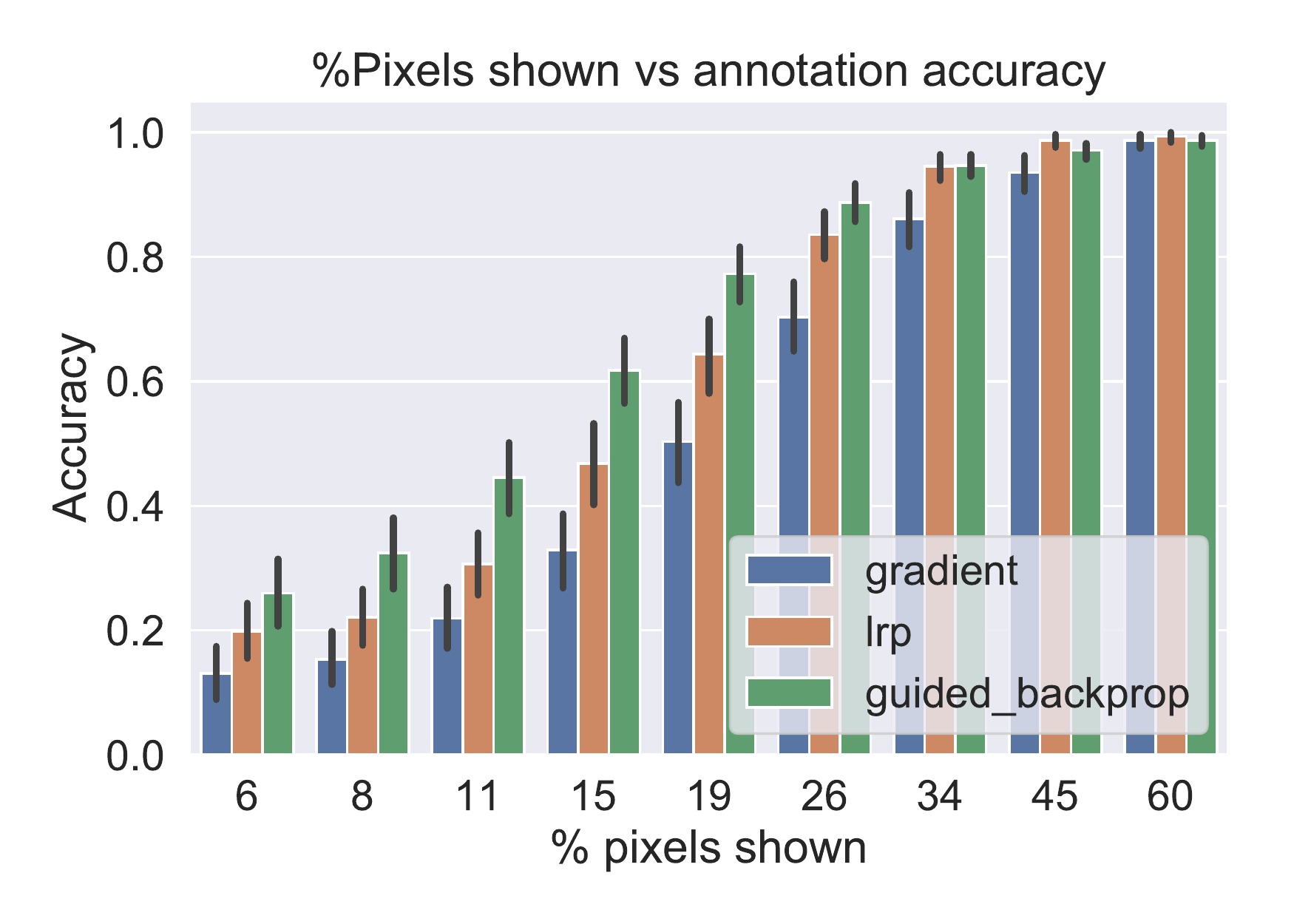}
\caption{Annotation accuracies of all subjects for different interpretability methods and mask sizes. For small mask sizes, the annotation accuracy is as low as 20\%; when 40\% of the pixels are masked, all subjects reliably detect the correct emotional expression. For intermediate levels of masking, there is a clear ranking of interpretability methods, guided backprop achieves highest accuracies.}
\label{fig:accuracy}
\end{figure}
\paragraph{Annotation accuracy distinguishes transparency methods}
The most important metric for our purposes is the annotation accuracy for different interpretability methods. Higher quality explanations should lead to higher annotation accuracy. Indeed we find that annotation accuracy clearly distinguishes the three interpretability methods used in our experiments. In \autoref{fig:accuracy} we show the annotation accuracy, averaged across subjects, for increasing mask sizes and all three different transparency approaches. Explanations using the plain gradient approach consistently led to the lowest annotation accuracy. The layerwise relevance propagation approach (LRP) \cite{Lapuschkin2017} yielded slightly better annotation accuracies and annotators assisted with the Guided BackProp explanations \cite{Springenberg2014} were consistently better than all other annotators. This effect was strongly dependent on the mask size and most pronounced for intermediate mask sizes around 15\% of pixels shown. When more than 45\% pixels were shown, subjects could detect the emotional expression reliably in all conditions. These results suggest that annotation accuracy in psychophysical experiments can serve as a robust quality indicator for interpretability methods.

\subsection{Metrics with no humans in the loop}
\label{sec:model_results}
\begin{figure}
\includegraphics[width=8.5cm]{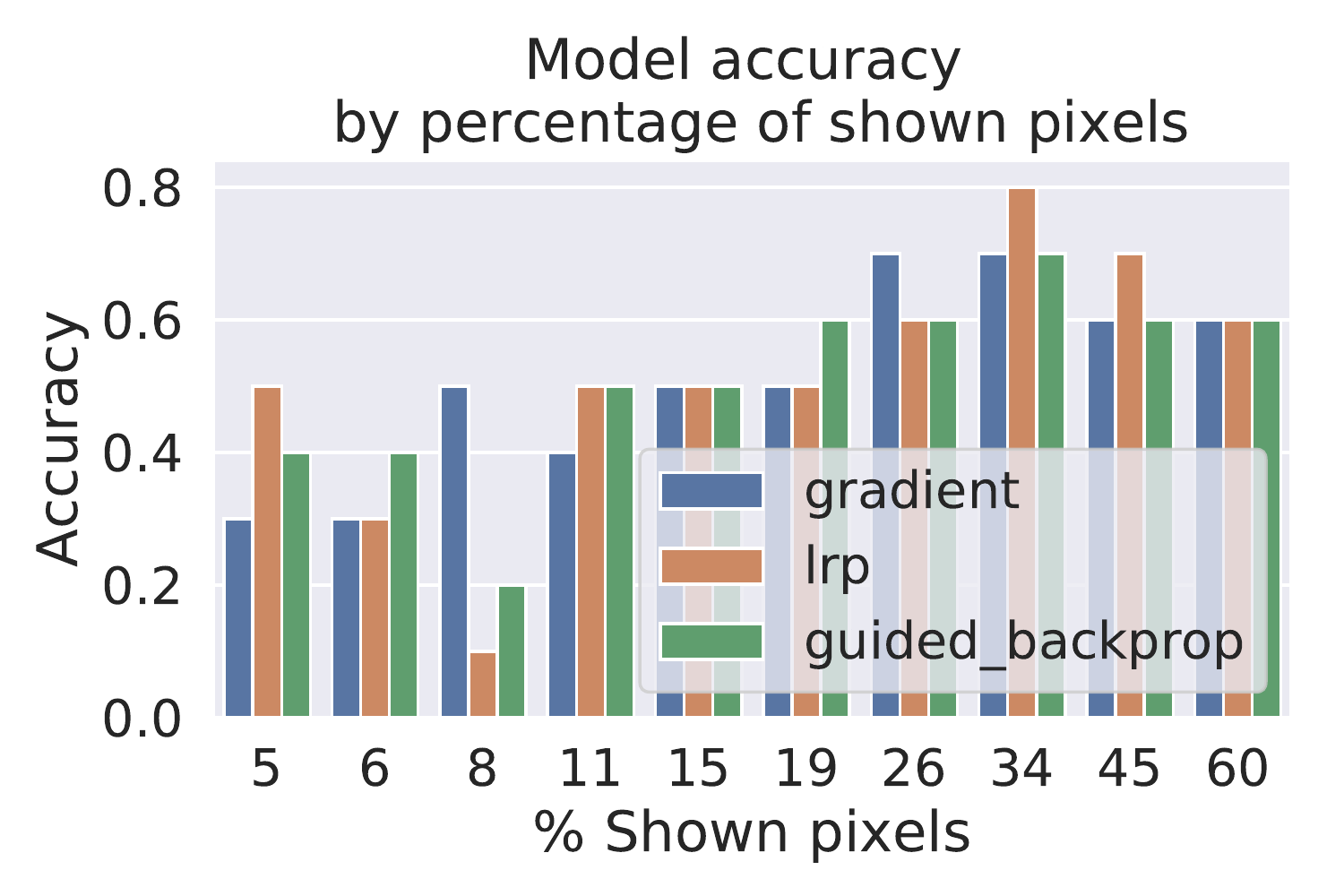}
\caption{Model prediction accuracies for different interpretability methods and mask sizes. In contrast to human annotators, there is no clear ranking of methods based on the accuracy of the ML predictions.}
\label{fig:accuracy_model}
\end{figure}

Next to the human in the loop experiments we also performed more standard experiments in which we tested the three interpretability approaches under the same experimental conditions as in the psychophysical experiments. For each mask size and interpretability method we computed the predictions of the EmoPy \cite{emopy} model and computed the accuracy across all images for each condition. The results are shown in \autoref{fig:accuracy_model} and demonstrate that at around 30\% of all pixels the model achieves the highest performance, that is better than the optimal performance on the test. When masking more pixels the prediction accuracy decreases irregularly without any specific trend, unlike in the case of the human annotators. An important difference to the human in the loop experiments is however that the model accuracy is not affected by the interpretability method as clearly as in the psychophysical experiments. Across all mask sizes there is no clear winner and in some cases the method that scored worst in the psychophysical experiments, Gradient, achieves the best accuracies when evaluating it on ML model predictions alone. 

Less important but interesting is also that while the general trend of lower accuracies with smaller masks is the same for both humans and machines in our experiments, human cognition tends to have a different sensitivity to the amount of pixels masked. While humans achieve a lower performance than the ML model when only 6\% of pixels are shown, this effect is reversed when more than 45\% of pixels are shown: annotators do not make mistakes in the emotional expression classification task  and the ML model only achieves accuracies around 60\% in those conditions. 

Both of these findings demonstrate that human cognition and machine cognition share some properties but are very different in others. In particular interpretability metrics that are purely based on machine predictions do not seem to capture what makes an explanation useful for humans. 

\begin{figure*}
\includegraphics[width=4.9cm]{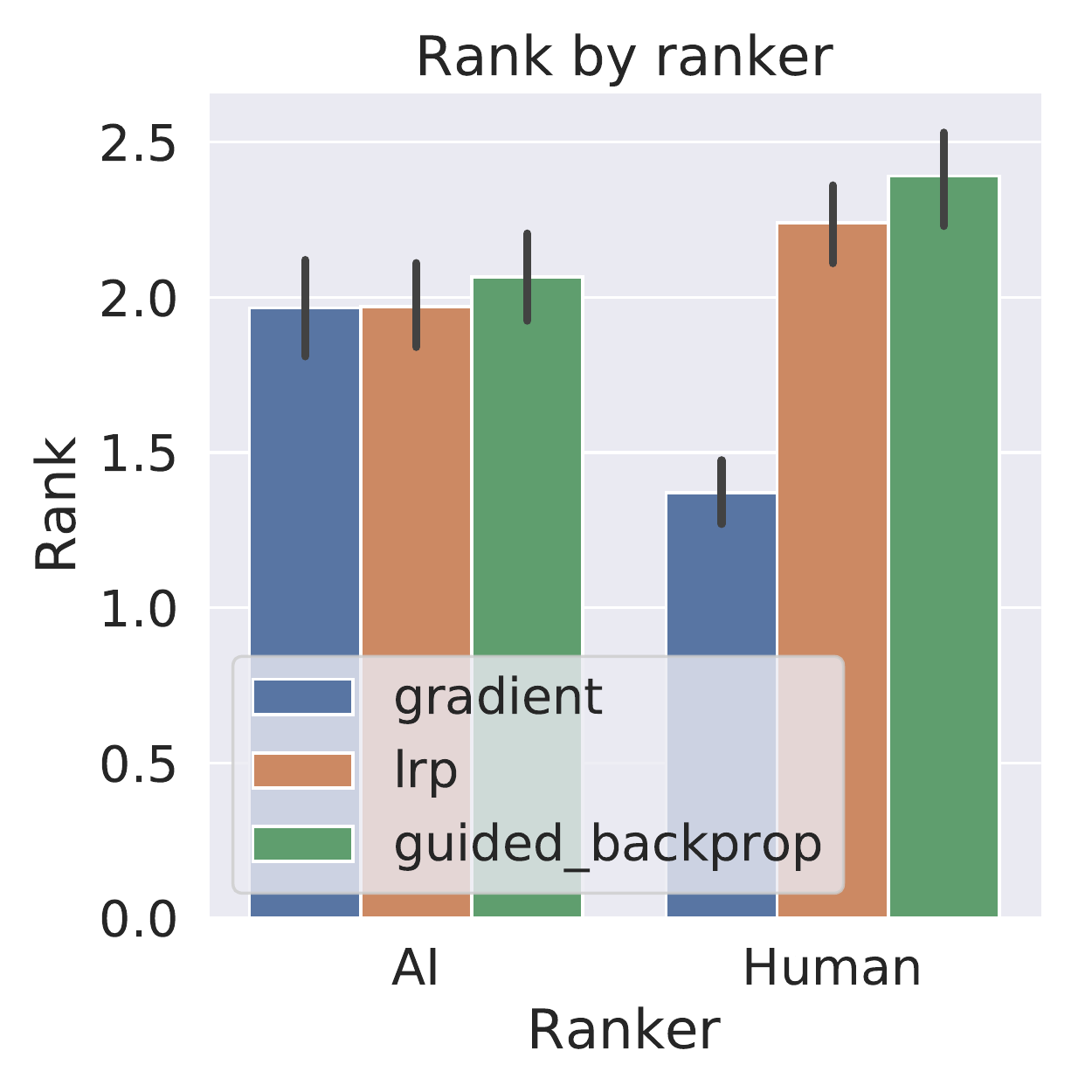}
\includegraphics[width=12.4cm]{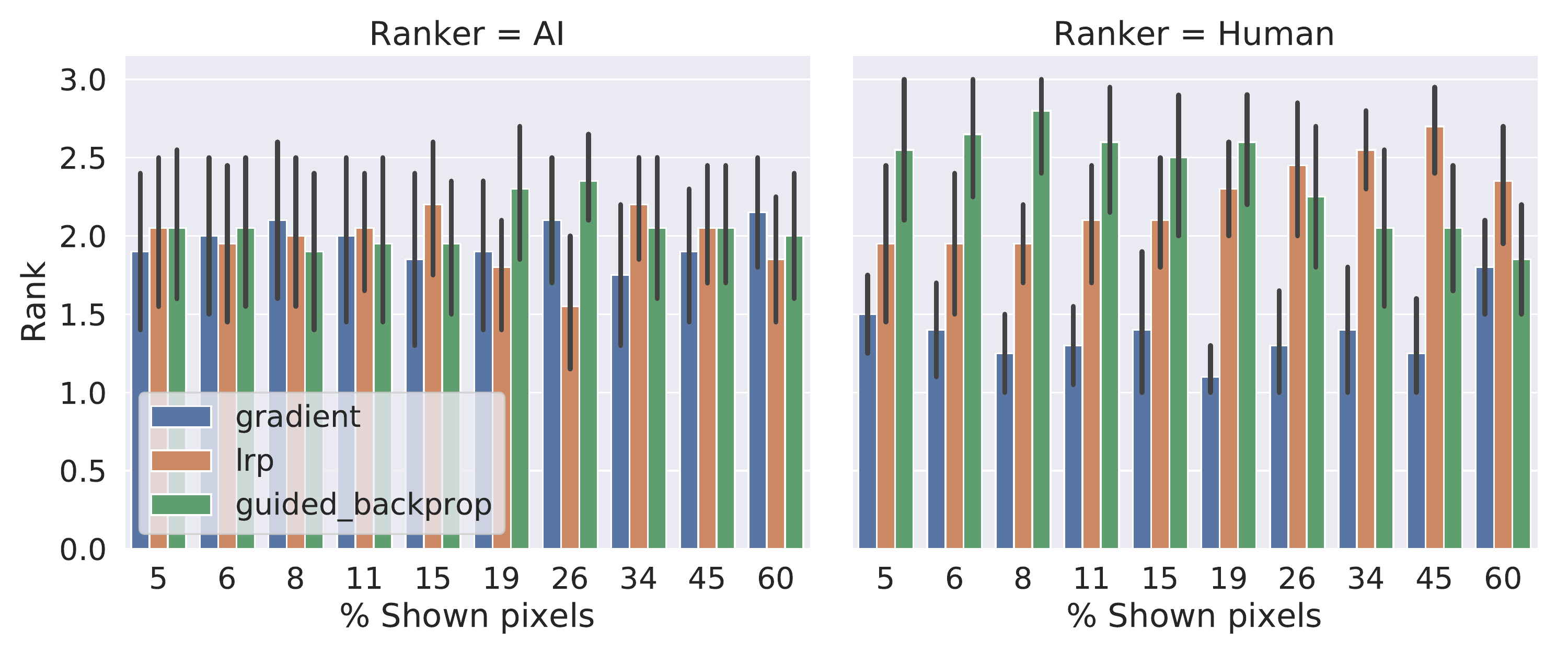}
\caption{Comparison of interpretability method rankings obtained in psychophysical experiments with human annotators and in experiments based on ML predictions with no humans in the loop. Ranks were computed for each image and then averaged across all images. For each image human ranks were based on annotation accuracy averaged across all subjects; AI ranks were based on cross-entropy loss per image. \textit{Left panel}: Rank for each interpretability method, averaged across all mask sizes. AI prediction based rankings show no clear differentiation of interpretability quality while rankings based on psychophysics show that Gradient based explanations are consistently worst and Guided BackProp explanations are consistently best. \textit{Middle and right panel}: Ranks for each interpretability method computed on AI predictions and human annotators for each mask size. For most mask sizes human annotators' accuracy was significantly higher for the Guided BackProp approach, there is no clear winner for the AI interpretability quality metric.}
\label{fig:rank_vs_threshold}
\end{figure*}

\subsection{Comparing NHIL and HIL metrics}
\label{sec:comparisons}
While the previous sections focussed on each metric individually we also compared the metrics obtained in psychophysical experiments with humans in the loop, see \autoref{sec:psychophysics_results}, with the metrics obtained by conventional offline no human in the loop (NHIL) approaches, see \autoref{sec:model_results}. For the comparison we paired the machine based NHIL metrics with the psychophysical human in the loop (HIL) metrics by grouping the data by image, mask size and interpretability method. For the HIL metrics we then computed the average accuracy across all subjects for each image and ranked all three interpretability methods according to the average annotation accuracy achieved with a given explanation assistance across all subjects for a given image. For the NHIL metrics we ranked the methods according to the cross-entropy loss incurred by a prediction for each interpretability method, mask size and image. As the cross-entropy loss is a continuous loss, in contrast to accuracy per data point, this allowed to rank the methods for each image despite the fact that there was only one prediction per image. The interpretability method rankings  for the psychophysics and machine based metrics are shown in \autoref{fig:rank_vs_threshold}. In the left panel the aggregated ranks across all mask sizes show that despite the transformation of the metrics into ranks, the two types of metrics are not very similar. Interpretability metrics obtained in the psychophysical experiments show a clear and robust ranking, while the rankings obtained by ML predictions do not allow to distinguish the three methods in terms of their interpretability. This is also reflected in the two right panels, which show the same data as in \autoref{fig:rank_vs_threshold}(\textit{left}), but split into all mask size conditions. The average ranks of the psychophysical experiments show the same clear pattern as the aggregate metrics, Guided BackProp is better than LRP which is in turn better than the plain Gradient explanation. In contrast it is difficult to single out the best interpretable explanation based on the machine based NHIL metrics; there is no significant difference between the methods for most thresholds, yet there seems to be a some advantage for Guided BackProp for mask sizes of 19\% and 26\%. Note that this trend is not reflected in the results of the psychophysical experiments. 

Overall these comparisons demonstrate that not only do machine based interpretability metrics not allow for a clear comparison of interpretability methods, more importantly these metrics are not representative of what is relevant for interpretabiltiy by humans either. 

\begin{figure}
\includegraphics[width=8.5cm]{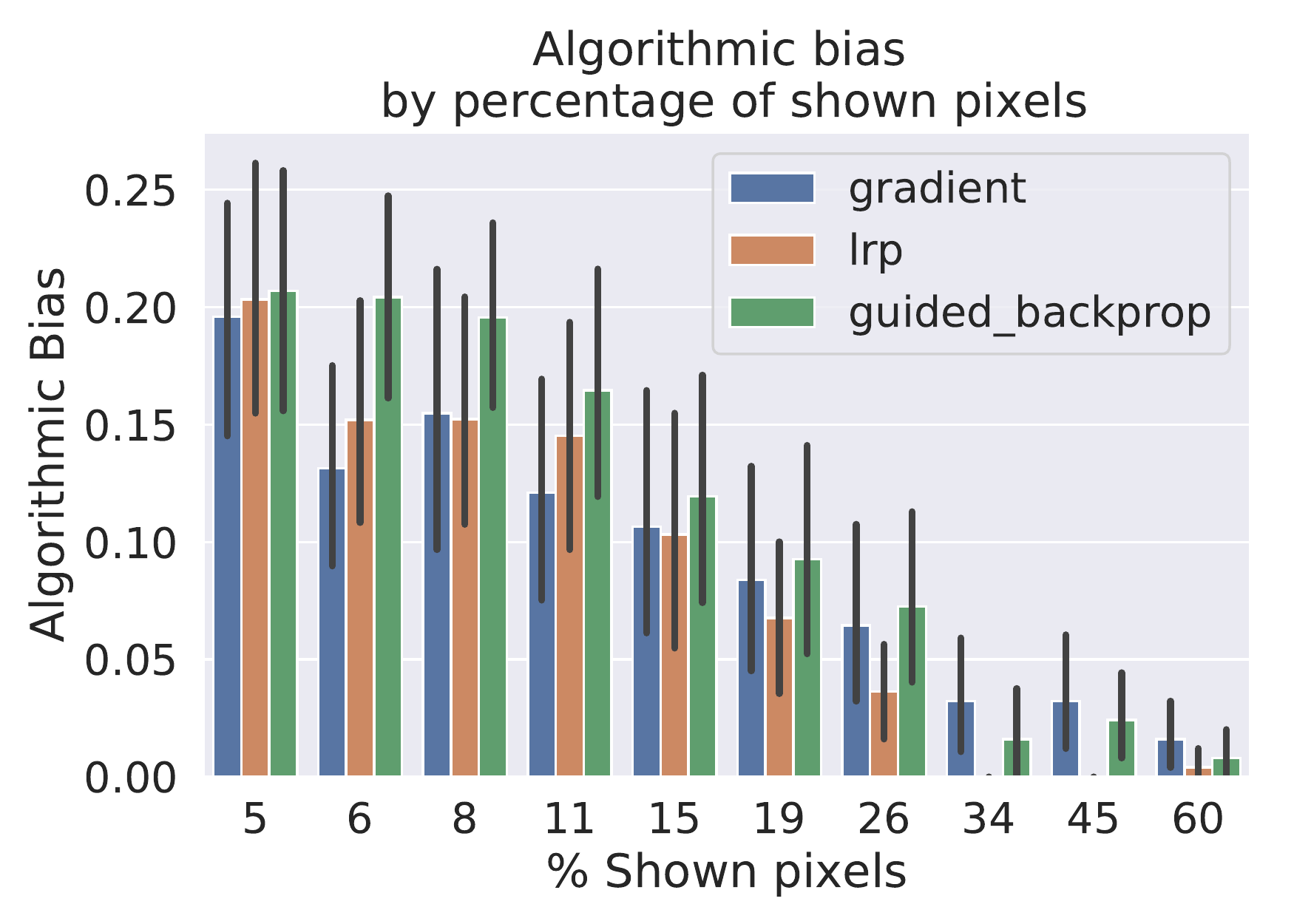}
\caption{Annotators' algorithmic bias measured as the overlap between human annotators and the ML prediction {\em when the ML prediction was incorrect}. The interpretability method Guided BackProp was most helpful in terms of annotation accuracy (see \autoref{fig:accuracy}), but at the same time it also appeared to lead annotators to follow the model prediction when it was wrong.}
\label{fig:algo_bias}
\end{figure}

\subsection{Transparency and algorithmic bias}
The above results demonstrate the impact of interpretability on annotation and prediction accuracy, but they miss an important aspect of transparent ML methods: human bias to algorithmic decisions. When explanations are intuitive humans tend to replicate the predictions of algorithms \cite{schmidt2019quantifying}, also in cases when the ML prediction is wrong. Such negative effects of transparency can be detrimental in real world applications. Hence measuring these effects helps to calibrate human-AI collaboration for more responsible and efficient usage of assistive AI technology. In \autoref{fig:algo_bias} we show the overlap of human annotators' predictions with the ML predictions, averaged across all images. Importantly we here only consider cases when the ML model was wrong, as we are interested in the negative aspects of algorithmic bias. For the mask sizes for which we see a clear advantage of the Guided BackProp method, from 6\% to 19\% percent pixels shown (\autoref{fig:rank_vs_threshold}), we also see a trend for larger algorithmic bias with explanations computed with the Guided BackProp method (\autoref{fig:algo_bias}). Annotators provided the same (wrong) answer as the model more often when they were exposed to the Gradient BackProp explanation compared to other explanations. This shows that more intuitive explanations not only lead to the increased annotation accuracy we have seen in \autoref{fig:rank_vs_threshold}, but also to the negative form of algorithmic bias when annotators wrongly replicate a model's prediction.

%% file: conclusion.tex

\section{Conclusion}
\label{sec:conclusion}

Methods that increase transparency of ML systems have become a major focus of research. Despite substantial advancements in the field and a plethora of methods available for rendering ML model predictions more interpretable, there appears to be no gold standard evaluation method for interpretability quality \cite{Guidotti2018}. Reliable and quantitative measures for evaluating interpretability are however a fundamental prerequisite for designing and improving transparent ML systems. 

Many studies use interpretability evaluations that rely on ML predictions only, without humans in the loop \cite{Samek2017}. This approach has the advantage that it is scalable and does not suffer from often subjective human judgements. But these measures are not directly related to the quantity of interest, how interpretable an explanation is for a human observer. Other studies evaluate interpretability in experiments with humans in the loop \cite{HUYSMANS2011,Lundberg2017,ribeiro2018anchors,lakkaraju2016interpretable,schmidt2019quantifying}. But these approaches do not follow the same experimental design which makes comparisons across studies difficult. To the best of our knowledge there are few studies that use the same experimental conditions for humans and machines when evaluating interpretability methods and that relate results from human in the loop experiments to evaluations without humans. 

In this study we used psychophysical experiments with humans to evaluate the quality of explanations for ML predictions. We compared those quality metrics with the metrics obtained in experiments without humans in the loop. Our results demonstrate that while psychophysical experiments allow to derive robust and clear rankings of interpretability quality, interpretability metrics obtained with ML predictions alone do not show a clear ranking of interpretability methods. More importantly our results also show that the metrics computed without humans in the loop are not only instable, they are also not representative of the rankings obtained in psychophysical experiments. 
These results highlight the potential of standardized psychophysical tests for the evaluation of ML methods and indicate that evaluations of interpretability should not rely exlusively on experiments without humans in the loop.

%% file: manuscript.bbl

\begin{thebibliography}{34}


\ifx \showCODEN    \undefined \def \showCODEN     #1{\unskip}     \fi
\ifx \showDOI      \undefined \def \showDOI       #1{#1}\fi
\ifx \showISBNx    \undefined \def \showISBNx     #1{\unskip}     \fi
\ifx \showISBNxiii \undefined \def \showISBNxiii  #1{\unskip}     \fi
\ifx \showISSN     \undefined \def \showISSN      #1{\unskip}     \fi
\ifx \showLCCN     \undefined \def \showLCCN      #1{\unskip}     \fi
\ifx \shownote     \undefined \def \shownote      #1{#1}          \fi
\ifx \showarticletitle \undefined \def \showarticletitle #1{#1}   \fi
\ifx \showURL      \undefined \def \showURL       {\relax}        \fi
\providecommand\bibfield[2]{#2}
\providecommand\bibinfo[2]{#2}
\providecommand\natexlab[1]{#1}
\providecommand\showeprint[2][]{arXiv:#2}

\bibitem[\protect\citeauthoryear{Alber, Lapuschkin, Seegerer, H{\"{a}}gele,
  Sch{\"{u}}tt, Montavon, Samek, M{\"{u}}ller, D{\"{a}}hne, and
  Kindermans}{Alber et~al\mbox{.}}{2018}]%
        {Alber2018}
\bibfield{author}{\bibinfo{person}{Maximilian Alber},
  \bibinfo{person}{Sebastian Lapuschkin}, \bibinfo{person}{Philipp Seegerer},
  \bibinfo{person}{Miriam H{\"{a}}gele}, \bibinfo{person}{Kristof~T.
  Sch{\"{u}}tt}, \bibinfo{person}{Gr{\'{e}}goire Montavon},
  \bibinfo{person}{Wojciech Samek}, \bibinfo{person}{Klaus-Robert
  M{\"{u}}ller}, \bibinfo{person}{Sven D{\"{a}}hne}, {and}
  \bibinfo{person}{Pieter-Jan Kindermans}.} \bibinfo{year}{2018}\natexlab{}.
\newblock \showarticletitle{{iNNvestigate neural networks!}}
\newblock  (\bibinfo{date}{aug} \bibinfo{year}{2018}).
\newblock
\showeprint[arxiv]{1808.04260}
\urldef\tempurl%
\url{http://arxiv.org/abs/1808.04260}
\showURL{%
\tempurl}


\bibitem[\protect\citeauthoryear{Cook}{Cook}{1977}]%
        {Cook1977}
\bibfield{author}{\bibinfo{person}{R.~Dennis Cook}.}
  \bibinfo{year}{1977}\natexlab{}.
\newblock \showarticletitle{Detection of Influential Observation in Linear
  Regression}.
\newblock \bibinfo{journal}{\emph{Technometrics}} \bibinfo{volume}{19},
  \bibinfo{number}{1} (\bibinfo{year}{1977}), \bibinfo{pages}{15--18}.
\newblock
\showISSN{00401706}
\urldef\tempurl%
\url{http://www.jstor.org/stable/1268249}
\showURL{%
\tempurl}


\bibitem[\protect\citeauthoryear{de~Leeuw}{de~Leeuw}{2015}]%
        {deleeuw2015}
\bibfield{author}{\bibinfo{person}{Joshua~R. de Leeuw}.}
  \bibinfo{year}{2015}\natexlab{}.
\newblock \showarticletitle{{jsPsych: A JavaScript library for creating
  behavioral experiments in a Web browser}}.
\newblock \bibinfo{journal}{\emph{Behavior Research Methods}}
  \bibinfo{volume}{47}, \bibinfo{number}{1} (\bibinfo{date}{mar}
  \bibinfo{year}{2015}), \bibinfo{pages}{1--12}.
\newblock
\showISSN{1554-3528}
\urldef\tempurl%
\url{https://doi.org/10.3758/s13428-014-0458-y}
\showDOI{\tempurl}


\bibitem[\protect\citeauthoryear{Deng, Dong, Socher, Li, Li, and Fei-Fei}{Deng
  et~al\mbox{.}}{2009}]%
        {imagenet_cvpr09}
\bibfield{author}{\bibinfo{person}{J. Deng}, \bibinfo{person}{W. Dong},
  \bibinfo{person}{R. Socher}, \bibinfo{person}{L.-J. Li}, \bibinfo{person}{K.
  Li}, {and} \bibinfo{person}{L. Fei-Fei}.} \bibinfo{year}{2009}\natexlab{}.
\newblock \showarticletitle{{ImageNet: A Large-Scale Hierarchical Image
  Database}}. In \bibinfo{booktitle}{\emph{CVPR09}}.
\newblock


\bibitem[\protect\citeauthoryear{Dietvorst, Simmons, and Massey}{Dietvorst
  et~al\mbox{.}}{2015}]%
        {Dietvorst2015}
\bibfield{author}{\bibinfo{person}{Berkeley~J. Dietvorst},
  \bibinfo{person}{Joseph~P. Simmons}, {and} \bibinfo{person}{Cade Massey}.}
  \bibinfo{year}{2015}\natexlab{}.
\newblock \showarticletitle{{Algorithm aversion: People erroneously avoid
  algorithms after seeing them err.}}
\newblock \bibinfo{journal}{\emph{Journal of Experimental Psychology: General}}
  \bibinfo{volume}{144}, \bibinfo{number}{1} (\bibinfo{date}{feb}
  \bibinfo{year}{2015}), \bibinfo{pages}{114--126}.
\newblock
\showISSN{1939-2222}
\urldef\tempurl%
\url{https://doi.org/10.1037/xge0000033}
\showDOI{\tempurl}


\bibitem[\protect\citeauthoryear{Doshi-Velez and Kim}{Doshi-Velez and
  Kim}{2017}]%
        {doshi2017towards}
\bibfield{author}{\bibinfo{person}{Finale Doshi-Velez} {and}
  \bibinfo{person}{Been Kim}.} \bibinfo{year}{2017}\natexlab{}.
\newblock \showarticletitle{Towards a rigorous science of interpretable machine
  learning}.
\newblock \bibinfo{journal}{\emph{arXiv preprint arXiv:1702.08608}}
  (\bibinfo{year}{2017}).
\newblock


\bibitem[\protect\citeauthoryear{EmoPy}{EmoPy}{[n. d.]}]%
        {emopy}
\bibfield{author}{\bibinfo{person}{EmoPy}.} \bibinfo{year}{[n. d.]}\natexlab{}.
\newblock \bibinfo{title}{\url{https://github.com/thoughtworksarts/EmoPy}}.
\newblock
\newblock


\bibitem[\protect\citeauthoryear{Fechner}{Fechner}{1860}]%
        {Fechner1860}
\bibfield{author}{\bibinfo{person}{Gustav~Theodor Fechner}.}
  \bibinfo{year}{1860}\natexlab{}.
\newblock \bibinfo{booktitle}{\emph{Elemente der Psychophysik}}.
\newblock \bibinfo{publisher}{Breitkopf und Hartel}.
\newblock


\bibitem[\protect\citeauthoryear{Guidotti, Monreale, Ruggieri, Turini,
  Giannotti, and Pedreschi}{Guidotti et~al\mbox{.}}{2018}]%
        {Guidotti2018}
\bibfield{author}{\bibinfo{person}{Riccardo Guidotti}, \bibinfo{person}{Anna
  Monreale}, \bibinfo{person}{Salvatore Ruggieri}, \bibinfo{person}{Franco
  Turini}, \bibinfo{person}{Fosca Giannotti}, {and} \bibinfo{person}{Dino
  Pedreschi}.} \bibinfo{year}{2018}\natexlab{}.
\newblock \showarticletitle{{A Survey of Methods for Explaining Black Box
  Models}}.
\newblock \bibinfo{journal}{\emph{Comput. Surveys}} \bibinfo{volume}{51},
  \bibinfo{number}{5} (\bibinfo{date}{aug} \bibinfo{year}{2018}),
  \bibinfo{pages}{1--42}.
\newblock
\showISSN{03600300}
\urldef\tempurl%
\url{https://doi.org/10.1145/3236009}
\showDOI{\tempurl}


\bibitem[\protect\citeauthoryear{Hajian, Bonchi, and Castillo}{Hajian
  et~al\mbox{.}}{2016}]%
        {Hajian2016}
\bibfield{author}{\bibinfo{person}{Sara Hajian}, \bibinfo{person}{Francesco
  Bonchi}, {and} \bibinfo{person}{Carlos Castillo}.}
  \bibinfo{year}{2016}\natexlab{}.
\newblock \showarticletitle{{Algorithmic Bias}}. In
  \bibinfo{booktitle}{\emph{Proceedings of the 22nd ACM SIGKDD International
  Conference on Knowledge Discovery and Data Mining - KDD '16}}.
  \bibinfo{publisher}{ACM Press}, \bibinfo{address}{New York, New York, USA},
  \bibinfo{pages}{2125--2126}.
\newblock
\showISBNx{9781450342322}
\urldef\tempurl%
\url{https://doi.org/10.1145/2939672.2945386}
\showDOI{\tempurl}


\bibitem[\protect\citeauthoryear{Hampel, Ronchetti, Rousseeuw, and
  Stahel}{Hampel et~al\mbox{.}}{2011}]%
        {hampel2011robust}
\bibfield{author}{\bibinfo{person}{Frank~R Hampel}, \bibinfo{person}{Elvezio~M
  Ronchetti}, \bibinfo{person}{Peter~J Rousseeuw}, {and}
  \bibinfo{person}{Werner~A Stahel}.} \bibinfo{year}{2011}\natexlab{}.
\newblock \bibinfo{booktitle}{\emph{Robust statistics: the approach based on
  influence functions}}. Vol.~\bibinfo{volume}{196}.
\newblock \bibinfo{publisher}{John Wiley \& Sons}.
\newblock


\bibitem[\protect\citeauthoryear{Haufe, Meinecke, G{\"o}rgen, D{\"a}hne,
  Haynes, Blankertz, and Bie{\ss}mann}{Haufe et~al\mbox{.}}{2014}]%
        {haufe2014interpretation}
\bibfield{author}{\bibinfo{person}{Stefan Haufe}, \bibinfo{person}{Frank
  Meinecke}, \bibinfo{person}{Kai G{\"o}rgen}, \bibinfo{person}{Sven
  D{\"a}hne}, \bibinfo{person}{John-Dylan Haynes}, \bibinfo{person}{Benjamin
  Blankertz}, {and} \bibinfo{person}{Felix Bie{\ss}mann}.}
  \bibinfo{year}{2014}\natexlab{}.
\newblock \showarticletitle{On the interpretation of weight vectors of linear
  models in multivariate neuroimaging}.
\newblock \bibinfo{journal}{\emph{Neuroimage}}  \bibinfo{volume}{87}
  (\bibinfo{year}{2014}), \bibinfo{pages}{96--110}.
\newblock


\bibitem[\protect\citeauthoryear{Herman}{Herman}{2017}]%
        {Herman2017ThePA}
\bibfield{author}{\bibinfo{person}{Bernease Herman}.}
  \bibinfo{year}{2017}\natexlab{}.
\newblock \showarticletitle{The Promise and Peril of Human Evaluation for Model
  Interpretability}.
\newblock \bibinfo{journal}{\emph{CoRR}}  \bibinfo{volume}{abs/1711.07414}
  (\bibinfo{year}{2017}).
\newblock


\bibitem[\protect\citeauthoryear{Huysmans, Dejaeger, Mues, Vanthienen, and
  Baesens}{Huysmans et~al\mbox{.}}{2011}]%
        {HUYSMANS2011}
\bibfield{author}{\bibinfo{person}{Johan Huysmans}, \bibinfo{person}{Karel
  Dejaeger}, \bibinfo{person}{Christophe Mues}, \bibinfo{person}{Jan
  Vanthienen}, {and} \bibinfo{person}{Bart Baesens}.}
  \bibinfo{year}{2011}\natexlab{}.
\newblock \showarticletitle{An empirical evaluation of the comprehensibility of
  decision table, tree and rule based predictive models}.
\newblock \bibinfo{journal}{\emph{Decision Support Systems}}
  \bibinfo{volume}{51}, \bibinfo{number}{1} (\bibinfo{year}{2011}),
  \bibinfo{pages}{141 -- 154}.
\newblock
\showISSN{0167-9236}
\urldef\tempurl%
\url{https://doi.org/10.1016/j.dss.2010.12.003}
\showDOI{\tempurl}


\bibitem[\protect\citeauthoryear{Kim}{Kim}{2015}]%
        {Kim2015}
\bibfield{author}{\bibinfo{person}{Been Kim}.} \bibinfo{year}{2015}\natexlab{}.
\newblock \emph{\bibinfo{title}{Interactive and interpretable machine learning
  models for human machine collaboration.}}
\newblock \bibinfo{thesistype}{Ph.D. Dissertation}.
  \bibinfo{school}{Massachusetts Institute of Technology}.
\newblock


\bibitem[\protect\citeauthoryear{Koh and Liang}{Koh and Liang}{2017}]%
        {pmlr-v70-koh17a}
\bibfield{author}{\bibinfo{person}{Pang~Wei Koh} {and} \bibinfo{person}{Percy
  Liang}.} \bibinfo{year}{2017}\natexlab{}.
\newblock \showarticletitle{Understanding Black-box Predictions via Influence
  Functions}. In \bibinfo{booktitle}{\emph{ICML}},
  \bibfield{editor}{\bibinfo{person}{Doina Precup} {and}
  \bibinfo{person}{Yee~Whye Teh}} (Eds.), Vol.~\bibinfo{volume}{70}.
  \bibinfo{pages}{1885--1894}.
\newblock
\urldef\tempurl%
\url{http://proceedings.mlr.press/v70/koh17a.html}
\showURL{%
\tempurl}


\bibitem[\protect\citeauthoryear{Lakkaraju, Bach, and Leskovec}{Lakkaraju
  et~al\mbox{.}}{2016}]%
        {lakkaraju2016interpretable}
\bibfield{author}{\bibinfo{person}{Himabindu Lakkaraju},
  \bibinfo{person}{Stephen~H Bach}, {and} \bibinfo{person}{Jure Leskovec}.}
  \bibinfo{year}{2016}\natexlab{}.
\newblock \showarticletitle{Interpretable decision sets: A joint framework for
  description and prediction}. In \bibinfo{booktitle}{\emph{Proceedings of the
  22nd ACM SIGKDD international conference on knowledge discovery and data
  mining}}. ACM, \bibinfo{pages}{1675--1684}.
\newblock


\bibitem[\protect\citeauthoryear{Lapuschkin, Binder, M{\"{u}}ller, and
  Samek}{Lapuschkin et~al\mbox{.}}{2017}]%
        {Lapuschkin2017}
\bibfield{author}{\bibinfo{person}{Sebastian Lapuschkin},
  \bibinfo{person}{Alexander Binder}, \bibinfo{person}{Klaus-Robert
  M{\"{u}}ller}, {and} \bibinfo{person}{Wojciech Samek}.}
  \bibinfo{year}{2017}\natexlab{}.
\newblock \showarticletitle{{Understanding and Comparing Deep Neural Networks
  for Age and Gender Classification}}.
\newblock  (\bibinfo{date}{aug} \bibinfo{year}{2017}).
\newblock
\showeprint[arxiv]{1708.07689}
\urldef\tempurl%
\url{http://arxiv.org/abs/1708.07689}
\showURL{%
\tempurl}


\bibitem[\protect\citeauthoryear{Lipton}{Lipton}{2016}]%
        {lipton2016mythos}
\bibfield{author}{\bibinfo{person}{Zachary~C Lipton}.}
  \bibinfo{year}{2016}\natexlab{}.
\newblock \showarticletitle{The mythos of model interpretability}.
\newblock \bibinfo{journal}{\emph{arXiv preprint arXiv:1606.03490}}
  (\bibinfo{year}{2016}).
\newblock


\bibitem[\protect\citeauthoryear{Lipton}{Lipton}{2017}]%
        {lipton2017doctor}
\bibfield{author}{\bibinfo{person}{Zachary~C Lipton}.}
  \bibinfo{year}{2017}\natexlab{}.
\newblock \showarticletitle{The Doctor Just Won't Accept That!}
\newblock \bibinfo{journal}{\emph{arXiv preprint arXiv:1711.08037}}
  (\bibinfo{year}{2017}).
\newblock


\bibitem[\protect\citeauthoryear{Lucey, Cohn, Kanade, Saragih, Ambadar, and
  Matthews}{Lucey et~al\mbox{.}}{2010}]%
        {Lucey2010}
\bibfield{author}{\bibinfo{person}{Patrick Lucey}, \bibinfo{person}{Jeffrey~F.
  Cohn}, \bibinfo{person}{Takeo Kanade}, \bibinfo{person}{Jason Saragih},
  \bibinfo{person}{Zara Ambadar}, {and} \bibinfo{person}{Iain Matthews}.}
  \bibinfo{year}{2010}\natexlab{}.
\newblock \showarticletitle{{The Extended Cohn-Kanade Dataset (CK+): A complete
  dataset for action unit and emotion-specified expression}}. In
  \bibinfo{booktitle}{\emph{2010 IEEE Computer Society Conference on Computer
  Vision and Pattern Recognition - Workshops}}. \bibinfo{publisher}{IEEE},
  \bibinfo{pages}{94--101}.
\newblock
\showISBNx{978-1-4244-7029-7}
\urldef\tempurl%
\url{https://doi.org/10.1109/CVPRW.2010.5543262}
\showDOI{\tempurl}


\bibitem[\protect\citeauthoryear{Lundberg and Lee}{Lundberg and Lee}{2017}]%
        {Lundberg2017}
\bibfield{author}{\bibinfo{person}{Scott~M. Lundberg} {and}
  \bibinfo{person}{Su{-}In Lee}.} \bibinfo{year}{2017}\natexlab{}.
\newblock \showarticletitle{A Unified Approach to Interpreting Model
  Predictions}. In \bibinfo{booktitle}{\emph{NIPS}}.
  \bibinfo{pages}{4768--4777}.
\newblock


\bibitem[\protect\citeauthoryear{Miller}{Miller}{2017}]%
        {miller2017explanation}
\bibfield{author}{\bibinfo{person}{Tim Miller}.}
  \bibinfo{year}{2017}\natexlab{}.
\newblock \showarticletitle{Explanation in artificial intelligence: insights
  from the social sciences}.
\newblock \bibinfo{journal}{\emph{arXiv preprint arXiv:1706.07269}}
  (\bibinfo{year}{2017}).
\newblock


\bibitem[\protect\citeauthoryear{Montavon, Lapuschkin, Binder, Samek, and
  M{\"{u}}ller}{Montavon et~al\mbox{.}}{2017}]%
        {Montavon2017}
\bibfield{author}{\bibinfo{person}{Gr{\'{e}}goire Montavon},
  \bibinfo{person}{Sebastian Lapuschkin}, \bibinfo{person}{Alexander Binder},
  \bibinfo{person}{Wojciech Samek}, {and} \bibinfo{person}{Klaus{-}Robert
  M{\"{u}}ller}.} \bibinfo{year}{2017}\natexlab{}.
\newblock \showarticletitle{Explaining nonlinear classification decisions with
  deep Taylor decomposition}.
\newblock \bibinfo{journal}{\emph{Pattern Recognition}}  \bibinfo{volume}{65}
  (\bibinfo{year}{2017}), \bibinfo{pages}{211--222}.
\newblock
\urldef\tempurl%
\url{https://doi.org/10.1016/j.patcog.2016.11.008}
\showDOI{\tempurl}


\bibitem[\protect\citeauthoryear{Ribeiro, Singh, and Guestrin}{Ribeiro
  et~al\mbox{.}}{2016}]%
        {lime}
\bibfield{author}{\bibinfo{person}{Marco~Tulio Ribeiro},
  \bibinfo{person}{Sameer Singh}, {and} \bibinfo{person}{Carlos Guestrin}.}
  \bibinfo{year}{2016}\natexlab{}.
\newblock \showarticletitle{"Why Should {I} Trust You?": Explaining the
  Predictions of Any Classifier}. In \bibinfo{booktitle}{\emph{{SIGKDD}}}.
  \bibinfo{pages}{1135--1144}.
\newblock


\bibitem[\protect\citeauthoryear{Ribeiro, Singh, and Guestrin}{Ribeiro
  et~al\mbox{.}}{2018}]%
        {ribeiro2018anchors}
\bibfield{author}{\bibinfo{person}{Marco~Tulio Ribeiro},
  \bibinfo{person}{Sameer Singh}, {and} \bibinfo{person}{Carlos Guestrin}.}
  \bibinfo{year}{2018}\natexlab{}.
\newblock \showarticletitle{Anchors: High-precision model-agnostic
  explanations}. In \bibinfo{booktitle}{\emph{AAAI Conference on Artificial
  Intelligence}}.
\newblock


\bibitem[\protect\citeauthoryear{Samek, Binder, Montavon, Lapuschkin, and
  M{\"{u}}ller}{Samek et~al\mbox{.}}{2017}]%
        {Samek2017}
\bibfield{author}{\bibinfo{person}{Wojciech Samek}, \bibinfo{person}{Alexander
  Binder}, \bibinfo{person}{Gr{\'{e}}goire Montavon},
  \bibinfo{person}{Sebastian Lapuschkin}, {and} \bibinfo{person}{Klaus{-}Robert
  M{\"{u}}ller}.} \bibinfo{year}{2017}\natexlab{}.
\newblock \showarticletitle{Evaluating the Visualization of What a Deep Neural
  Network Has Learned}.
\newblock \bibinfo{journal}{\emph{{IEEE} Trans. Neural Netw. Learning Syst.}}
  \bibinfo{volume}{28}, \bibinfo{number}{11} (\bibinfo{year}{2017}),
  \bibinfo{pages}{2660--2673}.
\newblock
\urldef\tempurl%
\url{https://doi.org/10.1109/TNNLS.2016.2599820}
\showDOI{\tempurl}


\bibitem[\protect\citeauthoryear{Schmidt and Biessmann}{Schmidt and
  Biessmann}{2019}]%
        {schmidt2019quantifying}
\bibfield{author}{\bibinfo{person}{Philipp Schmidt} {and}
  \bibinfo{person}{Felix Biessmann}.} \bibinfo{year}{2019}\natexlab{}.
\newblock \showarticletitle{Quantifying Interpretability and Trust in Machine
  Learning Systems}.
\newblock \bibinfo{journal}{\emph{arXiv preprint arXiv:1901.08558}}
  (\bibinfo{year}{2019}).
\newblock


\bibitem[\protect\citeauthoryear{Simonyan, Vedaldi, and Zisserman}{Simonyan
  et~al\mbox{.}}{2013}]%
        {Simonyan2013}
\bibfield{author}{\bibinfo{person}{Karen Simonyan}, \bibinfo{person}{Andrea
  Vedaldi}, {and} \bibinfo{person}{Andrew Zisserman}.}
  \bibinfo{year}{2013}\natexlab{}.
\newblock \showarticletitle{Deep Inside Convolutional Networks: Visualising
  Image Classification Models and Saliency Maps}.
\newblock \bibinfo{journal}{\emph{CoRR}}  \bibinfo{volume}{abs/1312.6034}
  (\bibinfo{year}{2013}).
\newblock
\showeprint[arxiv]{1312.6034}
\urldef\tempurl%
\url{http://arxiv.org/abs/1312.6034}
\showURL{%
\tempurl}


\bibitem[\protect\citeauthoryear{Sinha and Swearingen}{Sinha and
  Swearingen}{2002}]%
        {Sinha2002}
\bibfield{author}{\bibinfo{person}{Rashmi Sinha} {and} \bibinfo{person}{Kirsten
  Swearingen}.} \bibinfo{year}{2002}\natexlab{}.
\newblock \showarticletitle{{The role of transparency in recommender systems}}.
  In \bibinfo{booktitle}{\emph{CHI '02 extended abstracts on Human factors in
  computing systems - CHI '02}}. \bibinfo{publisher}{ACM Press},
  \bibinfo{address}{New York, New York, USA}, \bibinfo{pages}{830}.
\newblock
\showISBNx{1581134541}
\urldef\tempurl%
\url{https://doi.org/10.1145/506443.506619}
\showDOI{\tempurl}


\bibitem[\protect\citeauthoryear{Springenberg, Dosovitskiy, Brox, and
  Riedmiller}{Springenberg et~al\mbox{.}}{2014}]%
        {Springenberg2014}
\bibfield{author}{\bibinfo{person}{Jost~Tobias Springenberg},
  \bibinfo{person}{Alexey Dosovitskiy}, \bibinfo{person}{Thomas Brox}, {and}
  \bibinfo{person}{Martin Riedmiller}.} \bibinfo{year}{2014}\natexlab{}.
\newblock \showarticletitle{{Striving for Simplicity: The All Convolutional
  Net}}.
\newblock  (\bibinfo{date}{dec} \bibinfo{year}{2014}).
\newblock
\showeprint[arxiv]{1412.6806}
\urldef\tempurl%
\url{http://arxiv.org/abs/1412.6806}
\showURL{%
\tempurl}


\bibitem[\protect\citeauthoryear{Strumbelj and Kononenko}{Strumbelj and
  Kononenko}{2010}]%
        {Strumbel2010}
\bibfield{author}{\bibinfo{person}{Erik Strumbelj} {and} \bibinfo{person}{Igor
  Kononenko}.} \bibinfo{year}{2010}\natexlab{}.
\newblock \showarticletitle{An Efficient Explanation of Individual
  Classifications using Game Theory}.
\newblock \bibinfo{journal}{\emph{Journal of Machine Learning Research}}
  \bibinfo{volume}{11} (\bibinfo{year}{2010}), \bibinfo{pages}{1--18}.
\newblock
\urldef\tempurl%
\url{https://doi.org/10.1145/1756006.1756007}
\showDOI{\tempurl}


\bibitem[\protect\citeauthoryear{Zeiler and Fergus}{Zeiler and Fergus}{2014}]%
        {Zeiler2014}
\bibfield{author}{\bibinfo{person}{Matthew~D. Zeiler} {and}
  \bibinfo{person}{Rob Fergus}.} \bibinfo{year}{2014}\natexlab{}.
\newblock \showarticletitle{Visualizing and Understanding Convolutional
  Networks}. In \bibinfo{booktitle}{\emph{{ECCV}}}. \bibinfo{pages}{818--833}.
\newblock


\bibitem[\protect\citeauthoryear{Zien, Kr{\"{a}}mer, Sonnenburg, and
  R{\"{a}}tsch}{Zien et~al\mbox{.}}{2009}]%
        {Zien2009}
\bibfield{author}{\bibinfo{person}{Alexander Zien}, \bibinfo{person}{Nicole
  Kr{\"{a}}mer}, \bibinfo{person}{S{\"{o}}ren Sonnenburg}, {and}
  \bibinfo{person}{Gunnar R{\"{a}}tsch}.} \bibinfo{year}{2009}\natexlab{}.
\newblock \showarticletitle{The Feature Importance Ranking Measure}. In
  \bibinfo{booktitle}{\emph{{ECML} {PKDD} 2009}}. \bibinfo{pages}{694--709}.
\newblock


\end{thebibliography}
